\documentclass{article}

\PassOptionsToPackage{numbers, compress}{natbib}

\usepackage[dblblindworkshop, final]{neurips_2025}

\usepackage[utf8]{inputenc} 
\usepackage[T1]{fontenc}    
\usepackage{url}            
\usepackage{booktabs}       
\usepackage{amsfonts}       
\usepackage{nicefrac}       
\usepackage{microtype}      
\usepackage{xcolor}         
\usepackage{multirow}
\usepackage{graphicx}
\usepackage{glossaries}

\usepackage[capitalize,noabbrev,nameinlink]{cleveref}

\newacronym{llm}{LLM}{Large Language Model}
\newacronym{osm}{OSM}{OpenStreatMap}
\newacronym{poi}{POI}{Points of Interest}
\newacronym{nlu}{NLU}{Natural Language Understanding}
\newacronym{ret}{RET}{Rounting Engine Tool}
\newacronym{gct}{GCT}{Geospatial Context Tool}
\newacronym{crat}{CRAT}{Contextual Route Assessment Tool}
\newacronym{cot}{CoT}{Chain-of-Thought}
\newcommand{\NAME}{PAVe}

\title{Beyond Shortest Path: Agentic Vehicular Routing with Semantic Context}
\workshoptitle{UrbanAI: Harnessing Artificial Intelligence for Smart Cities}

\author{%
  Carnot Braun\thanks{Equal Contribution} \quad
  Rafael O. Jarczewski\footnotemark[1] \quad
  Gabriel U. Talasso\footnotemark[1] \\
  \textbf{Leandro A. Villas}\quad
  \textbf{Allan M. de Souza}\\
  Institute of Computing (IC) \\
  Hub de Inteligência Artificial e Arquiteturas Cognitivas (H.IAAC) \\
  Universidade Estadual de Campinas (UNICAMP) \\
  \texttt{\{c255785,r200219,g235078\}@dac.unicamp.br} \\
  \texttt{\{lvillas,allanms\}@unicamp.br} \\
}

\begin{document}

\maketitle

\begin{abstract}
Traditional vehicle routing systems efficiently optimize singular metrics like time or distance, and when considering multiple metrics, they need more processes to optimize . However, they lack the capability to interpret and integrate the complex, semantic, and dynamic contexts of human drivers, such as multi-step tasks, situational constraints, or urgent needs. This paper introduces and evaluates {\NAME} (Personalized Agentic Vehicular Routing), a hybrid agentic assistant designed to augment classical pathfinding algorithms with contextual reasoning.  Our approach employs a Large Language Model (LLM) agent that operates on a candidate set of routes generated by a multi-objective (time, CO$_2$) Dijkstra algorithm. The agent evaluates these options against user-provided tasks, preferences, and avoidance rules by leveraging a pre-processed geospatial cache of urban Points of Interest (POIs). In a benchmark of realistic urban scenarios, {\NAME} successfully used complex user intent into appropriate route modifications, achieving over 88\% accuracy in its initial route selections with a local model. We conclude that combining classical routing algorithms with an LLM-based semantic reasoning layer is a robust and effective approach for creating personalized, adaptive, and scalable solutions for urban mobility optimization.
\end{abstract}

\section{Introduction}
With the advance of Artificial Intelligent in urban environments, the best route recommendation problem moves from the chosen of fastest route between origin and destination to personalized paths based on different user's contexts and requirements that dynamically vary with safety concerns, environmental conditions, or personal preferences changing over time~\cite{PersonalizedRP,zhang2024surveyrouterecommendationsmethods,luca2021surveydeeplearninghuman}. Furthermore, route personalization enables intelligent traffic management and the optimization of urban resources, such as reducing congestion and pollutant emissions. However, much of this contextual and user preference information is descriptive, complex, and dynamic, making it difficult to model classical algorithmic solutions that does not deal with semantic contextual information. In this regard,~\gls{llm} emerges as an alternative for translating the semantic user's needs involved in describing preferences, taking into account the context and offering the best recommendations.

Recent works demonstrate the potential of~\gls{llm} and agent-based frameworks for diverse applications in the urban domain~\cite{wu2023smartagentbasedmodelinguse,xi2023risepotentiallargelanguage}. Applications vary from the construction of urban knowledge graphs and the simulation of multi-agent participatory planning to the analysis of mobility data to understand visit intent and the generation of personal mobility trajectories. These approaches open new avenues for the analysis and planning of smart cities~\cite{gong2024mobility, NEURIPS2024_e142fd2b, ning2024urbankgent}. Otherwise, in these works, it was observed the difficulty in making agents adapt to the different characteristics of the vehicular scenario, especially when trying to adjust to various types of~\gls{poi}, and since vehicle mobility data (routes, destinations,~\gls{poi}s related to vehicular services) are intrinsically linked to specific regions and their characteristics, making it challenging to apply a model trained in one domain to another without significant adaptations.

On the other hand, other solutions~\cite{qin2025lingotrip, FWTRoutes, WWIGN} use the reasoning capabilities of \gls{llm} to predict an individual's next destination on public transport, solving vehicle routing problems from natural language descriptions, and predict general patterns of human mobility. Such methods transform spatiotemporal data and problems into formats that \gls{llm} handle, allowing flexible solutions. However, these predictions are based on historical visits and may not accurately reflect the user's current desires or preferences in these contexts. This highlights the need for methods that consider both the user's context and the vehicular scenario.

As previous route recommendation systems do not take into account users' individual semantic intentions when suggesting routes~\cite{NEURIPS2024_e142fd2b}, in this article, we explore the potential of \gls{llm} agents for route personalization based on context and individual user needs. For this, we propose \NAME~(Personalized Agentic Vehicular Routing), an agent that plans and modifies routes prioritizing locations of interest and needs, taking into account characteristics such as urgency, preferences, time of day, and traffic. We opted for a hybrid approach to leverage the proven optimality and computational efficiency of classical algorithms for pathfinding, while dedicating the LLM's resources to its core strength: semantic reasoning and contextual understanding.
The main contributions of this paper can be summarized as follows: 

\begin{itemize}
    \item We develop {\NAME}, an agentic assistant for vehicular route planning and route selection, that considers contextual information, such as user route preferences and requirements, \gls{poi} and destination urgency and importance. 
    \item We evaluate {\NAME} into four realistic scenarios with real-world urban datasets with contextual street information and several user personalized requirements to fully assess the framework's capability and flexibility. 
    \item We made all prompts from different evaluation scenarios and user requests, as well as the code implementation of {\NAME} publicly available for future research and developments~\footnote{\url{https://github.com/carnotbraun/BsP_PAVe}.}.
\end{itemize}

\section{Personalized Agentic Vehicular Routing}

\begin{figure}[!ht]
    \centering
    \includegraphics[width=0.90\textwidth]{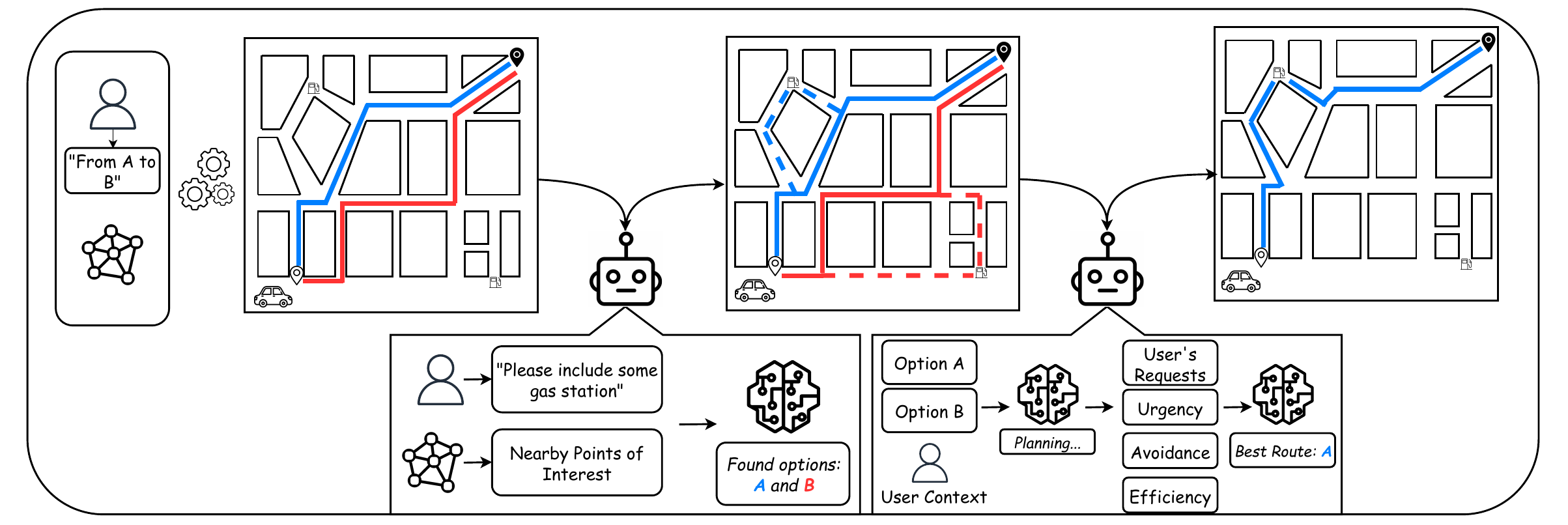}
    \caption{The overall of our \NAME~framework. \textbf{a)} Where the User chooses the desired trajectory \textbf{b)} With the information of the map, the path algorithm chooses the best two options. \textbf{c)} The LLM Agentic Assistant considers the context of the User and the Environment to include the POIs of the desire in the two options. \textbf{d)} The LLM translates all this for the path algorithm to recalculate the routes. \textbf{e)} The LLM chooses the best option for the User, considering the context, and returns the output.}
    \label{fig:solution}
\end{figure}

{\NAME} is a ReAct Agentic Assistant that integrates classical location and graph tools~\cref{fig:solution}. The methodology is designed to translate complex, contextual, multi-objective user requests into optimized, context-aware routing and decisions, focusing on efficiency, sustainability, personalization, and task completion.

First, user input is injected into the~\gls{llm}, which identifies and classifies the priority of each location within the user's request. We use~\gls{cot} for this,~\cite{wei2022chain}. This way, it can differentiate between an urgent and casual need in a request. Furthermore, our system is composed of three tools designed for robust decision-making based on users' preferences, described below.

\textbf{\gls{ret}.} Responsible for discovering the tuple of possible paths $\rho = (s,d)$, that is, user's origin $s$ and destination $d$. It operates on a graph-based representation of the urban road network $G = (V,E)$ where E and V represent, respectively, the sets of corners and streets on a city map. Generating a candidate set of paths $\mathcal{P} = \{\rho_0, \rho_1,\dots,\rho_k\}$ where $k$ is a set the routes from $s$ to $d$. The agent interacts with the map through this tool, finding the shortest paths to points of interest as well as extracting your characteristics.

\textbf{\gls{gct}.} This tool serves to ground the graph $G$ in the semantic context of the real world. The agent uses it to extract geospatial information, mainly \gls{poi}, such as supermarkets, hospitals, and parks along the routes encountered. For our experiments, we used information obtained from crowdsourced geographic data (\textbf{OpenStreetMap}). Its goal is to create a persistent link between these locations and their corresponding nodes in $G$, allowing the assistant agent to understand the environmental context of any route.

\textbf{\gls{crat}.} In this tool, the agent uses tools to identify the time of day, seasonal events, or periods of the year, such as school seasons, and, by cross-referencing this information, modify routes.

Thus,~\gls{llm} agent is responsible for making decisions and translating the semantic context of the user's goals, and building a route recommendation that meets the user's expectations.

\subsection{Agentic Operational Pipeline and Hierarchical Decision Making}

The pipeline consists of five steps:~\textbf{i) Task Classification}: The system receives the user's request in natural language. The~\gls{llm} classifies the task, assigning a priority (\texttt{URGENT} or \texttt{NORMAL}) and extracting relevant~\gls{osm} features (e.g., \texttt{{`amenity': `fuel'}}). \textbf{ii) Candidate Route Generation}: The~\gls{ret} is called and calculates a set of routes $P$ between the $s$ and $d$. Each candidate route is then annotated with metrics like total travel time and CO\textsubscript{2} emissions. \textbf{iii) Contextual Enrichment}: For each route, the system adds valuable layers of information using~\gls{osm}.
Then, if the task is classified as \texttt{URGENT}, an additional metric is calculated: the time needed to reach the nearest urgent~\gls{poi}. \textbf{iv) Agent Evaluation}: The~\gls{llm} receives a complete "dossier." This dossier contains detailed information about the candidate routes, including total time, CO\textsubscript{2} emissions, nearby~\gls{poi}, and, if applicable, the time to reach the urgent~\gls{poi}, along with the user's context and preferences using the~\gls{crat}. \textbf{v) Feedback Loop and Recalculation}: Based on the evaluation, the~\gls{llm} may request a new action, such as \texttt{ADD\_WAYPOINT}. When this occurs, the system initiates a feedback loop: a new route calculation is performed (Origin \textrightarrow{} Waypoint \textrightarrow{} Destination), dynamically adapting the final path.

\section{Experiments}

To evaluate the effectiveness of \NAME~, in this section, we propose four evaluation scenarios based on real user requirements and public urban maps of Luxembourg~\cite{codeca2017luxembourg}: Simple Task, Urgency, Avoidance, and Preferences. All scenarios were chosen after a search for routes that presented the characteristics we wanted to evaluate, varying route length and complexity. The evaluation showed that \NAME~ not only succeeds in choosing the best routes based on context and user preferences, but also prioritizes points of interest based on semantic information, such as urgency. In this paper we consider ``best route" like the shortest path from $s$ to the~\gls{poi} and the shortest path from~\gls{poi} to $d$. In addition to a comprehensive qualitative analysis of the results, we performed quantitative metrics, such as the accuracy of best route selection and task completion rate.

\subsection{Scenarios}
\label{sec:scenario}
To better evaluate \NAME~'s capabilities, we present three scenarios: \textbf{(a) The simple scenario}, which replicates more common and routine situations, such as going to the supermarket or passing through a park; \textbf{(b) Urgency}, which attempts to reflect scenarios in which passing through a given location is not optional, but rather necessary, such as stopping at a gas station with your car almost empty, since, even with two valid options, the best option would be to get to the point of interest faster; \textbf{(c) Avoidance}, which demonstrates scenarios in which the user's context and preference is to avoid a specific area; \textbf{(d) Efficiency}, which attempts to evaluate the choice of the route based on the distance, time and eco routing.

The experiments were conducted within a simulated urban environment, that is based on the \textbf{Luxembourg SUMO Traffic (LuST)} scenario, which provides a realistic road network forming the multi-weighted directed graph $G$, where the edge $E$ weights represent both estimated travel time and a CO\textsubscript{2} emission heuristic, which aim to suggest routes that are based on both the shortest distance and the most eco-friendly. To incorporate real-world context,~\gls{poi} were sourced from \textbf{\gls{osm}} using the \texttt{osmnx} library. This geospatial data was subsequently pre-processed into a local \texttt{GeoPackage} cache via \texttt{geopandas}, with each POI pre-linked to its nearest SUMO network node to enable efficient runtime queries. The decision-making agent is powered by a locally-hosted Qwen-series \gls{llm} from the \textbf{Hugging Face Hub} and also the API model(GPT-o3), executed via the \texttt{transformers} library. 

To understand the metrics we use, we first need to understand the difference between preferences and requirements. In this work, we use preferences as something that is not a priority, but rather desired (e.g., ``...I want to pass through a park on the way to the grocery store..."), and requirements as something we consider necessary to do, e.g., ``...I need to go to the gas station before going to the supermarket...". It is important to note that ~\gls{poi} is not necessarily a requirement, depending on~\gls{poi} (\texttt{UGERNT} or \texttt{NORMAL}). That way, \textbf{(i) Accuracy}: represents how many times the agent chose the best route (top-k) that best matches the user's preferences and requirements, and \textbf{(ii) Completeness}: represents how many times the agent successfully edited the route correctly to add~\gls{poi} based on user requests, i.e. even if~\gls{poi} is not \texttt{URGENT} the agent chose the path that passes through it.

\subsection{Main Results}

\begin{table}[ht]
\centering
\caption{Performance analysis of \NAME~across all simulated scenarios with different models using accuracy and completeness metrics (\textbf{Bold} are the best between them).}
\begin{tabular}{lcccc}
\toprule
\multirow{2}{*}{\textbf{Scenarios}} & \multicolumn{2}{c}{\textbf{Local} - \textit{Qwen 3 - 4B}} & \multicolumn{2}{c}{\textbf{API} - \textit{gpt-o3}} \\
\cmidrule(r){2-3} \cmidrule(r){4-5}
& Acc. (\%) & Complet. (\%) & Acc.(\%) & Complet.(\%) \\
\midrule
Simple        & \textbf{83.33} & 66.66 & 77.76 & 66.66  \\
Urgency       & 91.66 & 58.33 & \textbf{95.83} & 66.66  \\
Avoidance     & \textbf{75.0}  & 91.66 & 25.00 & 50.00  \\
Efficiency    & \textbf{93.30} & 95.66 & 88.88 & 88.88  \\
Overall       & \textbf{88.24} & 76.47 & 76.27 & 73.33 \\
\bottomrule
\end{tabular}
\label{tab:results}
\end{table}
The primary objective of this work was to explore the potential of a \gls{llm} agent not as a replacement for, but as an intelligent assistant to, state-of-the-art routing algorithms. Our proposed \NAME~ agent is designed to bridge the gap between human semantic intent and algorithmic execution by translating complex, context-dependent user tasks into structured data. To evaluate this methodology, agent performance was compared in four scenarios with four different paths each executed three times, comparing a smaller, locally-hosted model (\textit{Qwen 3 - 4B}) and also a larger, general-purpose API-based model (\textit{GPT-o3}). The results are summarized in Table \ref{tab:results}. Furthermore, we conducted some more experiments varying the number of routes considered (top-k), which can be seen in the appendix.

Overall, the local \textbf{Qwen model demonstrated superior performance}, achieving higher Accuracy (88.24\%) and Completeness (76.47\%) than its larger API-based counterpart. The most significant divergence occurred in the \textbf{Avoidance} scenarios, where the local model's ability to adhere to strict negative constraints was substantially better (75.0\% Acc. vs. 25.0\% Acc.). The results, particularly the consistently lower \textbf{Completeness} metric, also highlight two primary limitations of the current framework: an inconsistency in the agent's generation of the correct action schema, and a functional limitation wherein the feedback loop can only process a single waypoint addition.

These limitations provide a clear roadmap for future research. Future work will focus on enhancing action reliability through advanced prompting techniques like~\gls{cot} and fine-tuning the local model on a curated dataset of `(context, action)` pairs. Furthermore, a significant functional enhancement will be the integration of a multi-stop route planner. This would evolve the LLM's role from identifying a single waypoint to generating a \textit{list} of required waypoints, with the orchestrator then solving the resulting Traveling Salesperson Problem (TSP) variant to determine the optimal stop order. To make the agent truly personalized, we also plan to develop a user profile module that learns preferences and avoidance patterns over time, potentially using \textbf{Distributed Learning} to preserve user privacy. Finally, the system's sustainability metrics can be improved by integrating high-fidelity context from real-time traffic and vehicle-specific emission APIs, further grounding the agent's decisions in measurable, real-world impact.

\section*{Acknowledgments}
This project was supported by the brazilian Ministry of Science, Technology and Innovations, with resources from Law nº 8,248, of October 23, 1991, within the scope of PPI-SOFTEX, coordinated by Softex and published Arquitetura Cognitiva (Phase 3), DOU 01245.003479/2024-10.

\bibliographystyle{unsrtnat}
\bibliography{references}





\newpage
\appendix

\section*{Appendix}
\section{Related Works}
In this subsection, we gonna enter into more details about the works briefly discussed in the first section of this paper.

\subsection{LLMs for Urban Knowledge and Planning}

Recent works have explored the application of \gls{llm}s (LLMs) to various urban computing tasks. Some works present ~\cite{ning2024urbankgent} a unified LLM agent framework designed to automate Urban Knowledge Graph Construction (UrbanKGC), addressing the significant manual effort traditionally required. Their implementation involves fine-tuning Llama 2 and Llama 3 models on knowledgeable instruction sets derived from GPT-4. A key component is a tool-augmented iterative refinement module that invokes external geospatial tools to enhance the model's geospatial reasoning. The framework was evaluated on datasets from New York City and Chicago, focusing on Relational Triplet Extraction (RTE) and Knowledge Graph Completion (KGC) tasks. Performance was assessed using accuracy, measured through both human evaluation and GPT-4 self-evaluation. In a related domain~\cite{zhou2024largelanguagemodelparticipatory}, a multi-agent framework using LLMs to simulate participatory urban planning, aiming to generate land-use plans that reflect diverse resident needs efficiently. Their implementation uses `gpt-4-vision-preview' for a planner agent that interprets maps and `gpt-3.5-turbo-1106' for resident agents. To manage collaboration among a large number of agents, they employ a fishbowl discussion mechanism. The system was experimentally deployed in two Beijing urban regions, Huilongguan (HLG) and Dahongmen (DHM). Its effectiveness was measured against human experts and a Deep Reinforcement Learning baseline, using need-agnostic metrics like Service and Ecology, alongside novel need-aware metrics, Satisfaction and Inclusion.

\subsection{LLMs for Mobility and Trajectory Analysis}

The domain of human mobility has also seen an influx of LLM-based approaches. Recently, a team developed some very promising work that proposes~\cite{NEURIPS2024_e142fd2b} an agent framework for generating personal mobility trajectories by leveraging the semantic understanding of LLMs. The framework operates in two phases: first, it identifies habitual activity patterns using a self-consistency evaluation, and second, it performs motivation-driven activity generation through retrieval-augmented strategies. The implementation relies on GPT-o3 APIs and was validated on a real-world personal activity trajectory dataset from Tokyo. The evaluation compared the generated data distributions to real-world data using Jensen-Shannon Divergence (JSD) across four metrics: Step Distance (SD), Step Interval (SI), Daily Activity Routine Distribution (DARD), and Spatio-temporal Visits Distribution (STVD). Similarly, a unified framework for analyzing check-in sequences across multiple downstream tasks~\cite{gong2024mobility}. It aims to achieve a deeper semantic understanding of user behavior by capturing both visiting intentions and travel preferences. The implementation features a Visiting Intention Memory Network (VIMN) and a Human Travel Preference Prompt (HTPP) pool, with `TinyLlama-1B` as the backbone model, fine-tuned using LoRA. The framework was tested on four benchmark datasets, including Gowalla and Foursquare, for Next Location Prediction (LP), Trajectory User Link (TUL), and Time Prediction (TP) tasks. Evaluation metrics included Accuracy@k, Mean Reciprocal Rank (MRR), Mean Absolute Error (MAE), and Root Mean Squared Error (RMSE).

\subsection{LLMs for Routing and Trip Prediction}

Beyond analysis and generation, LLMs are being applied to complex routing and prediction problems. A model was developed that uses `ChatGPT-3.5-Turbo' in a zero-shot, in-context learning setting for predicting an individual's next trip~\cite{qin2025lingotrip}. Their implementation addresses the limitations of data-driven models by formatting historical mobility data into long-term, mid-term, and short-term contexts, which are fed to the LLM via a structured prompt designed to guide its reasoning. The model was evaluated on an Automated Fare Collection (AFC) dataset from the Hong Kong Mass Transit Railway (MTR). Its performance was benchmarked against traditional models using Accuracy and Weighted F1 score, particularly showing strength in few-shot learning scenarios. In the broader context of routing, the ability of LLMs to solve Vehicle Routing Problems (VRPs) directly from natural language descriptions~\cite{FWTRoutes}. They constructed a benchmark dataset of 21 VRP variants and evaluated the performance of GPT-4 and Gemini 1.0 Pro. Their proposed framework improves performance through a self-reflection mechanism that includes self-debugging of generated code and self-verification using LLM-generated unit tests. The experiments were evaluated on three primary metrics: feasibility (ratio of valid solutions), optimality (ratio of optimal solutions), and efficiency (normalized solution quality).

\section{PAVe Overview}

\begin{figure}[!ht]
    \centering
    \includegraphics[width=0.75\linewidth]{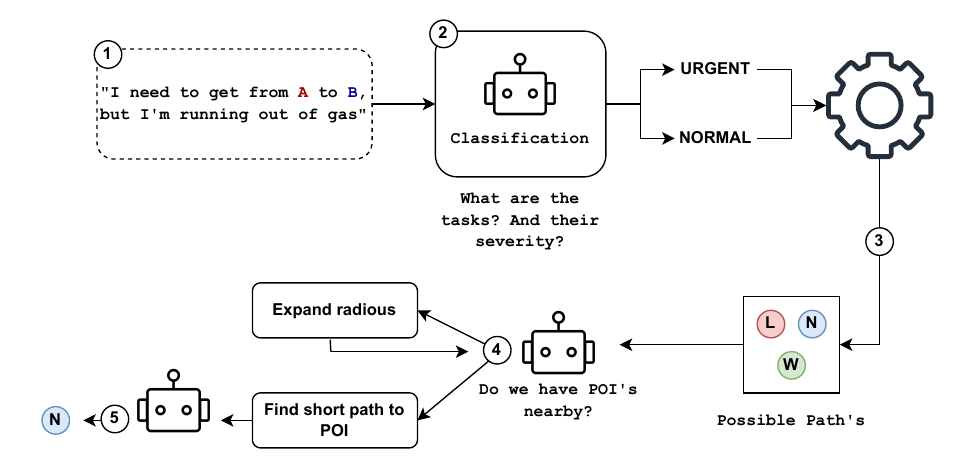}
    \caption{PAVe System Overview}
    \label{fig:agent_overview}
\end{figure}

The agent's decision-making process follows a structured, multi-stage pipeline, as illustrated in the main paper. This section provides a detailed technical description of each stage, as we can see in the image of the overview~\cref{fig:agent_overview}.

\begin{enumerate}
    \item \textbf{User Request Input and Parsing:} The pipeline is initiated with a user's natural language request, which includes an origin (A), a destination (B), and a set of tasks (e.g., \textit{"I'm running out of gas"}). This unstructured input is parsed by the central orchestrator to initialize the routing query.

    \item \textbf{LLM-based Task Classification:} The raw tasks are passed to the LLM agent for classification. Using a specifically engineered prompt, the agent analyzes the semantic content to determine the severity (\texttt{URGENT} or \texttt{NORMAL})  and extracts the corresponding OpenStreetMap (OSM) tags for each sub-task (e.g., \texttt{\{`amenity': `fuel'\}}). This step translates unstructured human intent into a machine-readable set of objectives and constraints.

    \item \textbf{Multi-Objective Candidate Path Generation:} The routing engine computes a set of \textit{k} diverse, candidate paths between the origin and the final destination. Each path is quantitatively annotated with a vector of objective metrics, including estimated travel time and a CO\textsubscript{2} emission heuristic, forming the basis for the subsequent evaluation.

    \item \textbf{Geospatial Contextual Enrichment:} Each candidate path undergoes an enrichment process. If an \texttt{URGENT} task was identified, an additional metric representing the time to reach the nearest relevant POI is calculated for each path. Concurrently, the \texttt{POIFinder} module performs a spatial query against a pre-processed geospatial cache to identify and attach a list of all normal-priority POIs that lie within a predefined radius of each route's geometry.

    \item \textbf{Hierarchical Evaluation and Action:} The enriched candidate routes, along with the full user and scenario context, are compiled into a final "dossier" for the LLM agent. The agent executes a hierarchical reasoning process to evaluate the options and select the optimal route that best satisfies the complete set of user needs. If this decision requires a route modification (e.g., adding a non-urgent waypoint), the agent's structured output triggers a feedback loop, where the orchestrator recalculates the final trajectory to include the new point of interest.
\end{enumerate}

\section{Other Results}

To evaluate the sensitivity of the ~\NAME~ agent to the number of initial choices, we conducted an experiment by varying the number of candidate paths (\textit{k}) generated by the Routing Engine. We compared the performance of three distinct \gls{llm} for \textit{k} values of 1, 3, 5, 10, and 20. The results, summarized in Table \ref{tab:k-table}, reveal significant differences in their decision-making capabilities and robustness as task complexity increases, considering that all the results consist of the overall scenario(The mean between all other~\cref{sec:scenario}).

The results highlight a critical trade-off between peak performance and stability. The \texttt{Qwen3-4B} model achieved the highest accuracy (88.24\%) for low \textit{k} values (1 and 3), but its performance degraded sharply as more options were introduced, with accuracy dropping to 54.90\% for \textit{k}=20. In contrast, the OpenAI API model, while not reaching the same initial peak, demonstrated remarkable stability. Its accuracy remained constant at 76.47\% for \textit{k} up to 10, with only a marginal decrease at \textit{k}=20. A similar pattern of high consistency was observed for its completeness metric. The \texttt{Phi-3.5} model consistently underperformed across all configurations.

This performance degradation, most prominent in the local models, is likely due to two factors. First, a larger \textit{k} increases the cognitive load on the LLM. Second, as \textit{k} increases, the marginal differences between candidate paths diminish, making the choice more ambiguous. This suggests that the OpenAI API model is more robust in handling such ambiguity.

\renewcommand{\arraystretch}{1.5}
\begin{table}[ht]
\centering
\caption{Performance analysis of PAVe on different top-k paths, and different models, using accuracy and completeness metrics.}
\label{tab:k-table}
\begin{tabular}{ccccccc}
\toprule
\multirow{2}{*}{\textbf{Routes}} &
  \multicolumn{2}{c}{\textbf{\begin{tabular}[c]{@{}c@{}}Qwen3\\ 4B\end{tabular}}} &
  \multicolumn{2}{c}{\textbf{\begin{tabular}[c]{@{}c@{}}Phi-3.5\\ 3.82B\end{tabular}}} &
  \multicolumn{2}{c}{\textbf{\begin{tabular}[c]{@{}c@{}}OpenAi - API\\ o3\end{tabular}}} \\ \cmidrule{2-7} 
              & \textbf{Acc. (\%)} & \textbf{Complet. (\%)} & \textbf{Acc. (\%)} & \textbf{Complet. (\%)} & \textbf{Acc. (\%)} & \textbf{Complet. (\%)} \\ \bottomrule
\textbf{k=1}  & \textbf{88.24}     & \textbf{76.47}         & 35.57              & 71.88                  & 76.47              & 73.33                  \\ 
\textbf{k=3}  & \textbf{88.24}     & \textbf{76.47}         & 33.33              & 68.75                  & 76.47              & 73.33                  \\ 
\textbf{k=5}  & \textbf{76.47}     & \textbf{73.33}         & 31.37              & 70.59                  & \textbf{76.47}     & \textbf{73.33}         \\ 
\textbf{k=10} & 66.67              & \textbf{73.33}         & 25.49              & 71.22                  & \textbf{76.47}     & \textbf{73.33}         \\ 
\textbf{k=20} & 54.90              & 64.70                  & 15.69              & 67.65                  & \textbf{74.51}     & \textbf{72.73}         \\ \bottomrule
\end{tabular}
\end{table}

The main conclusion is that for an agent like ~\NAME~, simply increasing the number of options is detrimental. Our findings suggest that future work should focus on a dynamic \textit{k}, potentially adapted to the number of points of interest (POIs) or the semantic diversity of the routes. Furthermore, integrating tools to pre-filter candidate routes could empower the agent to make more informed decisions from a high-quality, curated set of options.

\section{Prompts}

The~\NAME~agent's cognitive abilities are orchestrated through a series of structured prompts designed to elicit specific, reliable, and machine-parsable responses. We developed three distinct prompts for the primary stages of the agent's operational pipeline.

\begin{figure}[!ht]
    \centering
    \includegraphics[width=1.0\linewidth]{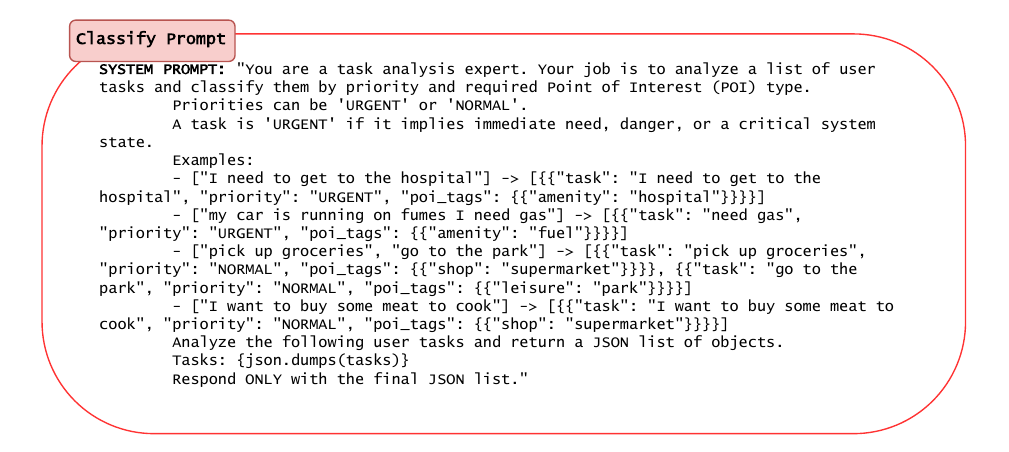}
    \caption{Classification Prompt}
    \label{fig:classify}
\end{figure}

\paragraph{Task Classification Prompt.}
This initial prompt~\cref{fig:classify} functions as the primary~\gls{nlu} component of the system. It is responsible for deconstructing a user's free-form request into a structured set of objectives. 
\textit{Input:} The prompt receives a list of user tasks as raw strings.
\textit{Output:} It returns a JSON list of objects, where each object represents a distinct sub-task and is annotated with a classified priority (\texttt{URGENT} or \texttt{NORMAL}) and a dictionary of the corresponding OpenStreetMap (OSM) tags required to locate a relevant Point of Interest (POI). This structured output serves as the foundational input for all subsequent planning and routing stages.

\begin{figure}[!ht]
    \centering
    \includegraphics[width=1.0\linewidth]{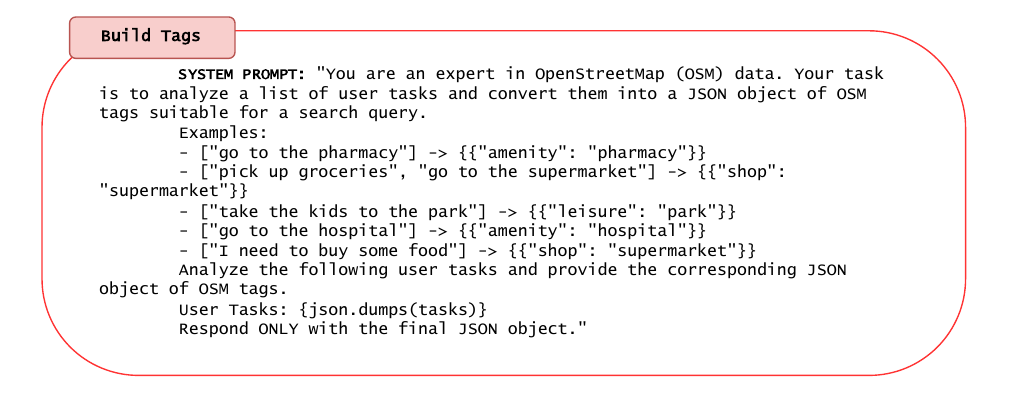}
    \caption{Builds Prompt}
    \label{fig:build}
\end{figure}

\begin{figure}[!ht]
    \centering
    \includegraphics[width=1.0\linewidth]{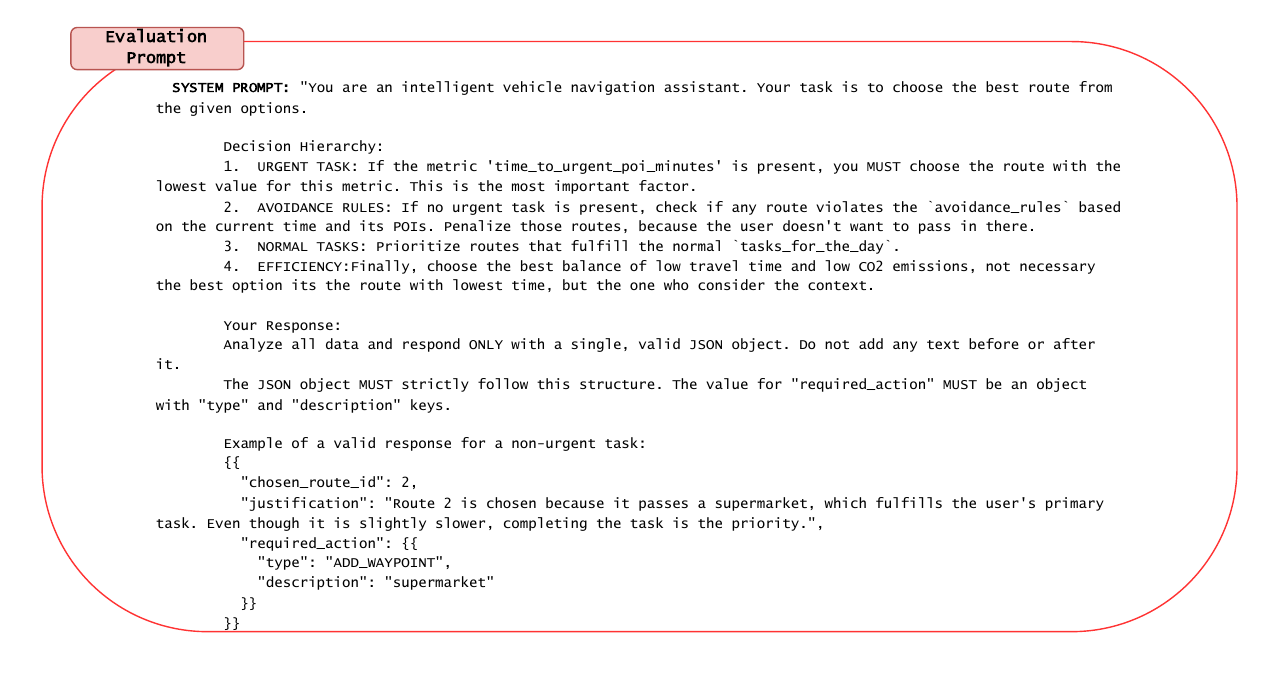}
    \caption{Evaluation Prompt}
    \label{fig:eval}
\end{figure}

\paragraph{Tag Aggregation Prompt.}
To prepare for geospatial queries, this utility prompt~\cref{fig:build} is used to consolidate the objectives from all classified tasks into a single, unified search query. 
\textit{Input:} It takes a list of user tasks as strings.
\textit{Output:} The prompt yields a single JSON object that aggregates all necessary OSM tags from the various user tasks, creating a master list of all POI types that need to be searched for within the given scenario.
\paragraph{Route Evaluation Prompt.}
This is the core reasoning prompt, where the agent performs its final, multi-objective decision-making~\cref{fig:eval}. It is explicitly instructed to follow a strict decision hierarchy to ensure predictable and logical behavior.
\textit{Input:} The prompt is provided with a comprehensive "dossier" in JSON format, containing the full user context (preferences, avoidance rules), the scenario context (time, traffic), and a list of candidate routes, each annotated with its respective metrics (e.g., travel time, CO\textsubscript{2} emissions, nearby POIs, and time to an urgent POI).
\textit{Output:} The agent returns a single, structured JSON object containing the ID of the chosen route, a detailed natural language justification for its decision, and a \texttt{required\_action} object that instructs the orchestrator on the next step, such as initiating a feedback loop to add a waypoint.

\end{document}